\documentclass[journal]{IEEEtran}
\usepackage{graphicx}
\usepackage{subfigure}
\usepackage{amssymb}
\usepackage{booktabs}
\usepackage{ textcomp }
\usepackage{color}
\usepackage{makecell}
\usepackage{multirow}
\usepackage{amsmath}
\usepackage{algorithmic}
\usepackage{cite}

\ifCLASSINFOpdf
\else
\fi
\begin{document}
%
\title{Lightweight Feature Fusion Network for\\Single Image Super-Resolution}
%
%
%

\author{Wenming~Yang, Wei~Wang, Xuechen~Zhang, Shuifa~Sun, Qingmin~Liao
\thanks{W.~Yang, W.~Wang, X.~Zhang and Q.~Liao are with the Department of Electronic Engineering, Graduate School at Shenzhen, Tsinghua University, China (E-mail: \{yang.wenming@sz, wangwei17@mails, xc-zhang16@mails, liaoqm@\}.tsinghua.edu.cn.}
\thanks {S.~Sun is with the Department of Hubei Key Laboratory of Intelligent Vision Based Monitoring for Hydroelectric Engineering, China Three Gorges University, China (E-mail: watersun@ctgu.edu.cn).}
}

%
%

\markboth{}%
{Yang \MakeLowercase{\textit{et al.}}: Lightweight Feature Fusion Network for Single Image Super-Resolution}
%



\maketitle

\begin{abstract}
Single image super-resolution(SISR) has witnessed great progress as convolutional neural network(CNN) gets deeper and wider. However, enormous parameters hinder its application to real world problems. In this letter, We propose a lightweight feature fusion network (LFFN) that can fully explore multi-scale contextual information and greatly reduce network parameters while maximizing SISR results. LFFN is built on spindle blocks and a softmax feature fusion module (SFFM). Specifically, a spindle block is composed of a dimension extension unit, a feature exploration unit and a feature refinement unit. The dimension extension layer expands low dimension to high dimension and implicitly learns the feature maps which is suitable for the next unit. The feature exploration unit performs linear and nonlinear feature exploration aimed at different feature maps. The feature refinement layer is used to fuse and refine features. SFFM fuses the features from different modules in a self-adaptive learning manner with softmax function, making full use of hierarchical information with a small amount of parameter cost. Both qualitative and quantitative experiments on benchmark datasets show that LFFN achieves favorable performance against state-of-the-art methods with similar parameters. Code is avaliable at https://github.com/qibao77/LFFN-master.
\end{abstract}


%
\IEEEpeerreviewmaketitle

\section{Introduction}
%
%
%
%
\IEEEPARstart{S}{ingle} image super-resolution (SISR) is an important low-level computer vision task which aims at recovering a high-resolution (HR) image from a low-resolution (LR) image. It is a seriously ill-posed problem since an LR image can be mapped to an infinite number of HR images. Recently, deep convolutional neural network (CNN) has greatly facilitated improvements in this field. Dong \emph{et~al.} \cite{dong2014learning} firstly proposed a three-layer CNN to establish a mapping between LR and HR. Kim \emph{et~al.} proposed the well-known VDSR \cite{Kim2016Accurate}, which introduced residual learning and adaptive gradient clipping to alleviate the difficulty of training deep network. In DRCN\cite{kim2016deeply}, the recursive network was used to reduce the model parameters and a multi-supervised strategy was adopted to fuse intermediate results. 
Benefiting from skip-connection can alleviate the vanishing-gradient problem \cite{he2016deep}, \cite{chu2018novel}, Lim \emph{et~al.} \cite{lim2017enhanced} built a very deep network MDSR (more than 160 layers) with residual blocks. 

\begin{figure}[ht]
\centering
\includegraphics[scale=0.5]{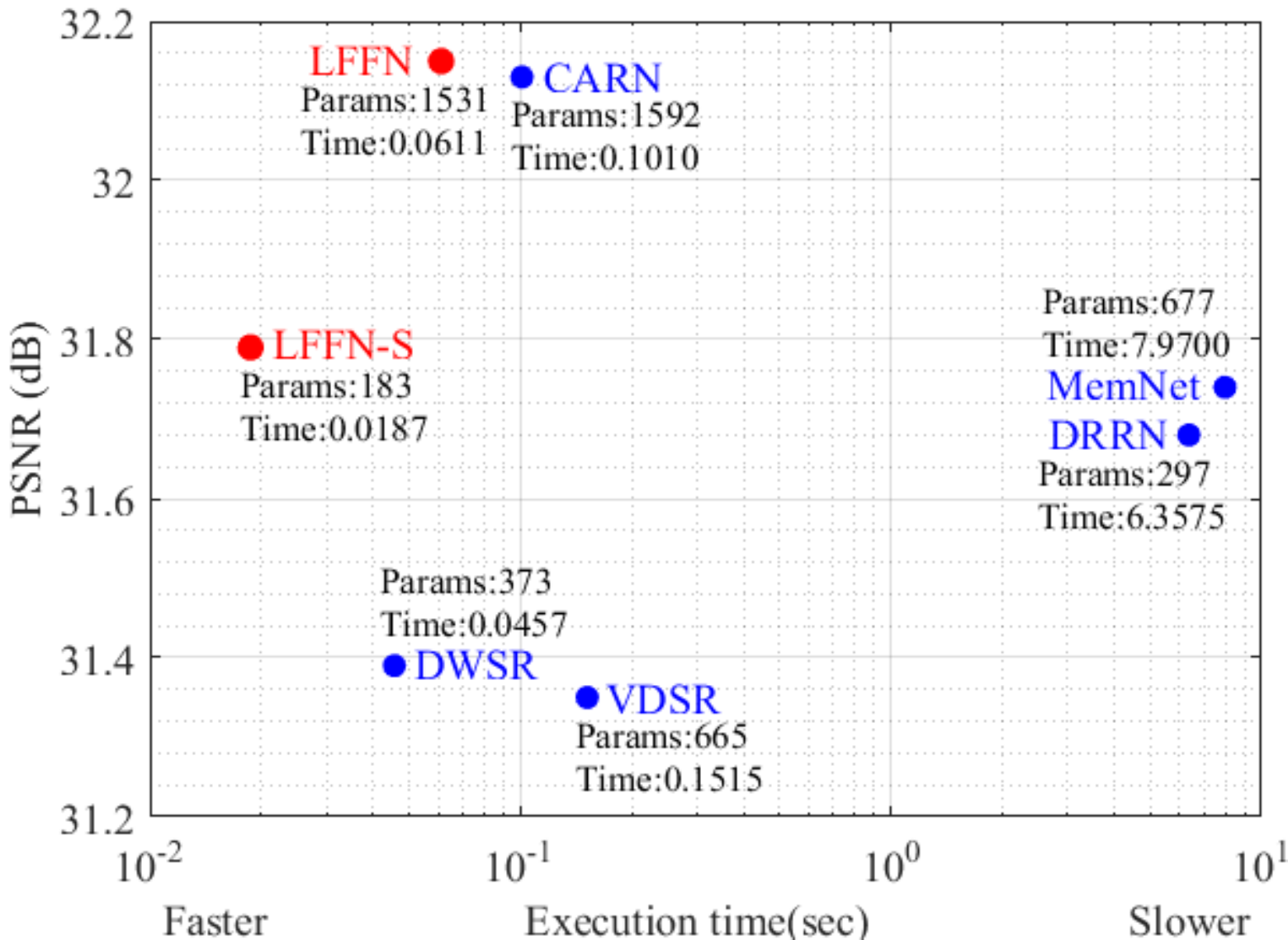}
\caption{
Speed and accuracy trade-off. The average PSNR and the average inference time for upscaling $\times 4$ on Set5. LFFN and LFFN-S contain B4M15 and B4M4, respectively. 
The 3*3 convolution of all spindle blocks in LFFN-S is replaced by depthwise convolution.}
\label{fig:trade_off_time_psnr}
\end{figure}

\par Researchers usually deepen and widen the network to achieve better performance. However, even constructed with small convolution kernels, such as $3 \times 3$, the network will take up large memories. 
In order to lighten the deep network, some strategies have been adopted. DRRN \cite{tai2017image} employed parameter sharing strategy to reduce parameters, but it still requires large computation objectively. CARN-M \cite{ahn2018fast} adopt group convolution to attack a trade-off between computation and performance of the model. Unfortunately, applying group convolution directly to SISR will obviously impair performance. To address these problems, we propose a lightweight network LFFN to compute the HR image from the original LR image. 
In LFFN, we introduce a new organization of the inception-residual block \cite{szegedy2017inception}, named spindle block, which contains a dimension extension unit, a feature exploration unit and a feature refinement unit. The dimension extension unit can learn the feature maps suitable for the next unit, and the architecture can be mitigated by fewer filters in backbone. Inspired by ResNeXt \cite{xie2017aggregated} and Xception \cite{chollet2017xception}, we introduce a feature exploration unit to explore the linear and nonlinear as well as multi-scale information for 4 different channel groups. This unit can improve the representational power of the model and can further alleviate the architecture due to fewer filters in each group. We also consider using feature maps of different receptive fields to enhance the performance. Taking computation into account and motivated by feature recalibration demonstrated in SENets \cite{hu2017squeeze}, we develop a softmax feature fusion module (SFFM) to aggregate the features of different levels in a self-adaptive channel-wise convex weighted way rather than the multi-supervised method used in DRCN\cite{kim2016deeply} and MemNet \cite{tai2017memnet}. The parameters of SFFM are not large, since there is only one dense layer applied to each global feature of different levels. And SFFM can learn how to combine the features that are most conducive to reconstruction.
\section{Proposed Method}

\begin{figure}[ht]
\centering
\includegraphics[scale=0.4]{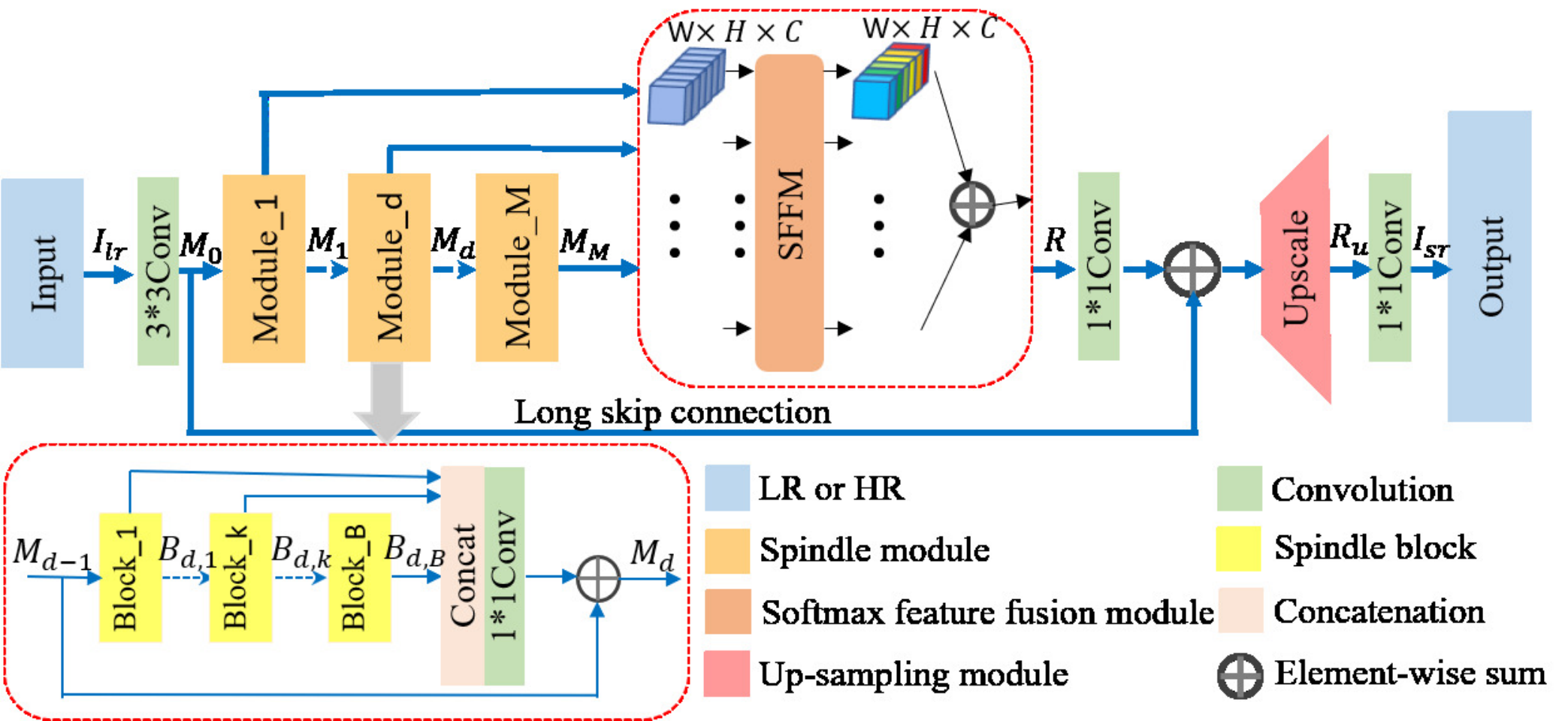}
\caption{The architecture of our proposed lightweight feature fusion network (LFFN). 
}
\label{fig:framework}
\end{figure}
\begin{figure}[ht]
\centering
\subfigure[ ]{
\begin{minipage}[t]{0.3\linewidth}
\centering
\includegraphics[scale=0.5]{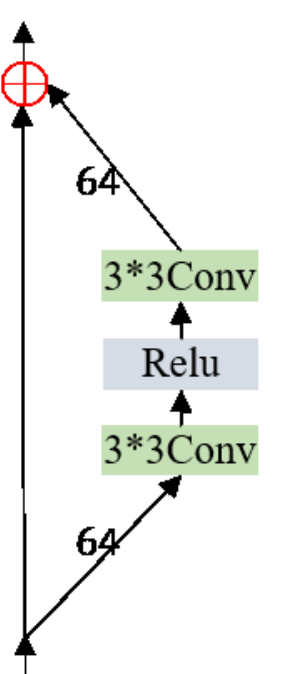}
\end{minipage}%
}%
\subfigure[ ]{
\begin{minipage}[t]{0.4\linewidth}
\centering
\includegraphics[scale=0.5]{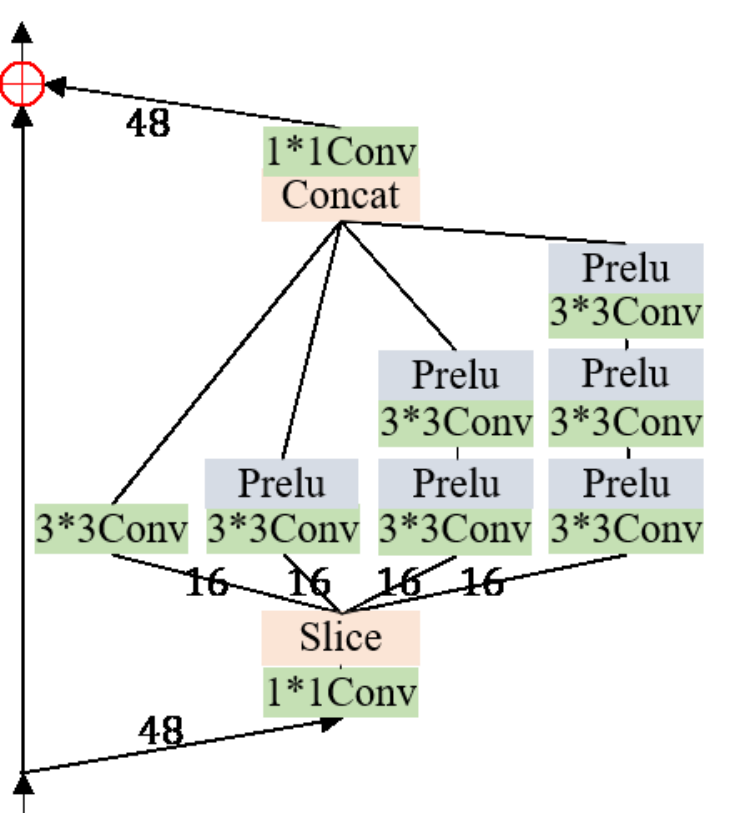}
\end{minipage}%
}%
\centering
\caption{Basic block of different architecture. (a) residual block. (b) spindle block. }
\label{basic_block}
\end{figure}

\subsection{Network structure}
As shown in Fig. \ref{fig:framework}, the overall architecture consists of $B\times M$ spindle blocks, a softmax feature fusion module (SFFM) and an up-sampling module. We denote $I_{lr}$ and $I_{sr}$ as the input and output of LFFN, respectively. First of all, we use a single $3\times 3$ convolutional layer of 48 filters to extract the feature maps from the original LR image:
\begin{equation}
M_{0}=F_{sf}(I_{lr}),
\label{first_layer}
\end{equation}
where $F_{sf}(\cdot)$ represents convolution operation and $M_{0}$ serves as the input of next part. The next part is $M$ stacked local feature fusion modules. Inspired by MemNet \cite{tai2017memnet} and SRDenseNet \cite{tong2017image}, we concatenate feature maps from $B$ stacked spindle blocks to further make full use of local features. We also introduce residual learning for each module to make deep network training easier. This procedure can be expressed as
\begin{equation}
\begin{aligned}
M_{d}&=F_{d}(M_{d-1})\\
&=F_{df}([B_{d,1},\cdot \cdot \cdot ,B_{d,k},\cdot \cdot \cdot ,B_{d,B}])+M_{d-1},
\end{aligned}
\label{module}
\end{equation}
where $F_{d}$ denotes the $d$-th module function and $F_{df}$ is the function of the $1\times 1$ convolution in $d$-th module. $M_{d}$ and $B_{d,k}$ indicate the output of the $d$-th module and $k$-th spindle block respectively. More details about spindle block will be given in next section. After extracting complicated features progressively with $M$ modules, we further conduct softmax feature fusion (SFFM).
\begin{equation}
R=F_{sffm}(M_{d},\cdot \cdot \cdot ,M_{d},\cdot \cdot \cdot ,M_{d}),
\label{softmax_1}
\end{equation}
where $R$ is the output feature maps of SFFM, $F_{sffm}$ denotes a composite function. 
Finally, like\cite{ledig2017photo} and \cite{lim2017enhanced} , we utilize ESPCN \cite{shi2016real} followed by a convolution layer to upscale the refined feature maps and get the output of LFFN. It is worth mentioning that we replace the $3\times 3$ convolution with $1\times 1$ convolution in upscale module and the last layer to further reduce parameters.
\begin{equation}
I_{sr}=F_{last}(F_{up}(F_{fuse}(R)+M_{0})),
\label{result_equation}
\end{equation}
where $F_{last}$ and $F_{fuse}$ denote the $1\times 1$ convolution and $F_{up}$ is the function of upscale module.
\subsection{Spindle Block}
To reap the benefits of inception residual block \cite{szegedy2017inception} and group convolution, we propose a well-designed residual block, named spindle block. The overall block can be formulated as:
\begin{equation}
B_{d,k}=B_{d,k-1}+F_{fru}(F_{feu}(F_{deu}(B_{d,k-1})),
\label{spindle_equation}
\end{equation}
where $F_{fru}$, $F_{feu}$ and $F_{deu}$ represent compound function of three basic units respectively. And more details about them explicated as follows.
\subsubsection{dimension extension unit}
The number of filters is a critical factor to improve the efficiency of deep networks, which is fixed to 64 in many deep methods for SISR currently. We can lighten the architecture by decreasing the filters, but the performance fluctuates accordingly. 
Using ``bottleneck layer'' \cite{szegedy2015going} ($1\times 1$ convolution) to compress dimensions resemble pooling operation in channel dimension. We believe that reducing feature channels before non-linear layer can lead to information loss. Here, we expand the dimensions from 48 to 64 before non-linear mapping to maintain performance with fewer parameters. 
\subsubsection{feature exploration unit}
As shown in Fig.\ref{basic_block}(b), we first slice the feature maps into four different 16-dimensional groups. Then, we explore nonlinear information in three groups and linear information in the other group. Specifically, we adopt a sequence of $3\times 3$ convolutional layers followed by parametric rectified linear units (PReLUs) to make full use of the image multi-scale information. Different from \cite{szegedy2015going}, \cite{szegedy2017inception}, we assemble linear and nonlinear information to boost representational power of basic blocks and directly dispose the expanded feature maps instead of reducing dimension by additional $1\times 1$ convolutions.
\subsubsection{feature refinement unit}
Then the concatenate feature maps are sent to a $1\times 1$ convolutional layer which acts as refining features, compressing dimensions and overcoming the impact of the slice operation on weakening the information flow.

\par Basically, as shown in Fig.\ref{basic_block}, our spindle block can take advantage of linear and nonlinear and multi-scale information with fewer parameters than baseline residual block. 
In particular, when we use the configuration expressed in Fig.\ref{basic_block}, a spindle block has $30.21 \%$ of parameters of a residual block. This ratio can be further decreased to $12.13 \%$ by replacing $3\times 3$ convolution in spindle block with depthwise convolution. 
More analysis will be described in experiment.
\subsection{Softmax Feature Fusion Module}
\begin{figure}[b]
\centering
\includegraphics[scale=0.5]{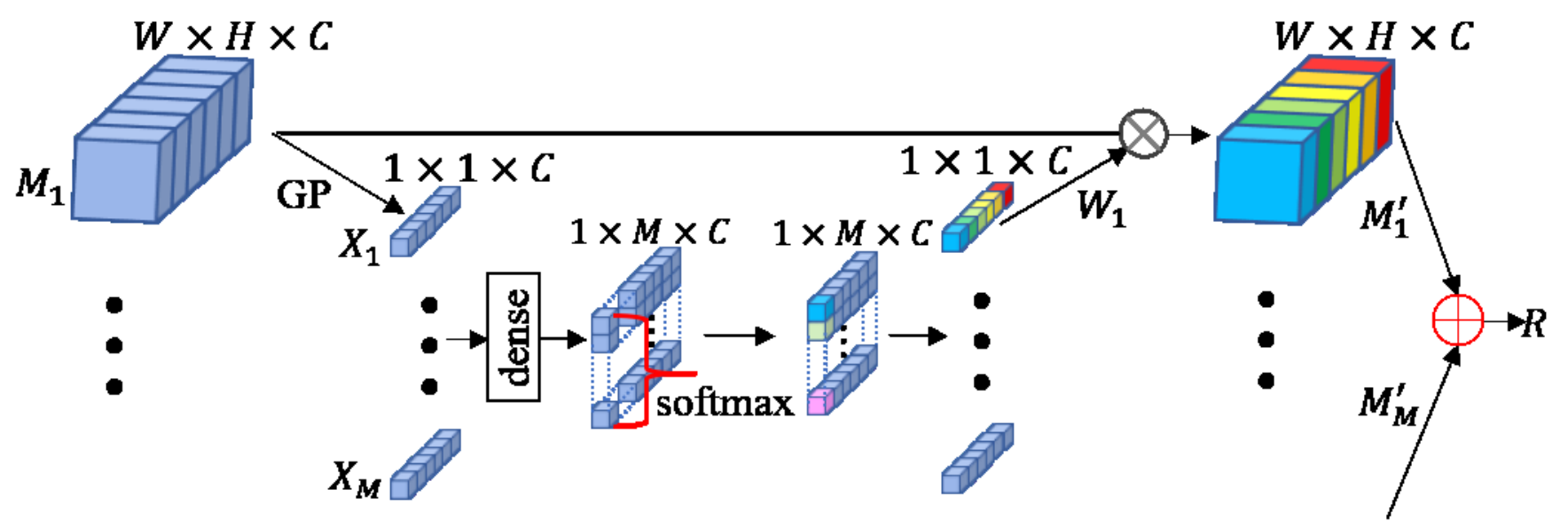}
\caption{Softmax feature fusion module (SFFM) architecture. $\otimes$ denotes element-wise product. GP represents global average pooling. ``dense'' represents the fully connected layers applied to global features from different modules.}
\label{fig:SFFM}
\end{figure}

Information in different levels of feature maps can complement each other for reconstruction. In order to gain more abundant and efficient information, we focus on hierarchical features and achieve a fusion mechanism. As shown in Fig.\ref{fig:SFFM}, we take all intermediate feature maps $M_{i}$ as input and generate a fusion representation $R$. And $M_{i}=\left [ m_{i1}, ..., m_{ij}, ...,m_{iC}  \right ]$, $i=1, ..., M$, $j=1, ..., C$, where $m_{ij}\in \mathbb{R}^{W\times H}$ denotes the $j$-th channel of the $i$-th feature maps $M_{i}$, and $C$ is the total number of channels. Inspired by squeeze operation in\cite{hu2017squeeze}, we apply global average pooling to each channel to obtain the global channel feature $X_{i}=\left [ x_{i1}, ..., x_{ij}, ..., x_{iC}  \right ], X_{i}\in  \mathbb{R}^{C}$. Then, we follow it with a dense layer to fully exploit inter-channel correlation, as formulated below:
\begin{equation}
Y_{i}=\alpha_{i}X_{i},
\label{dense_layer}
\end{equation}
where $\alpha_{i}$ represent the weight set of $i$-th dense layer and $Y_{i}=\left [ y_{i1}, ..., y_{ij}, ..., y_{iC} \right ]$, $Y_{i}\in \mathbb{R}^{C}$. We utilize concatenation and slice operation and softmax function to produce the weight of the corresponding channel of different features. This process can be expressed as:
\begin{equation}
W_{j}=softmax(Y_{j}),
\label{softmax_function}
\end{equation}
where $Y_{j}=\left [ y_{1j}, ..., y_{ij}, ..., y_{Mj}  \right ]$, $Y_{j}\in \mathbb{R}^{M}$ and  $W_{j}=\left [ w_{1j}, ..., w_{ij}, ..., w_{Mj}  \right ]$, $W_{j}\in \mathbb{R}^{M}$. The final output of SFFM is obtained as the following formula:
\begin{equation}
\begin{aligned}
r_{j}=\sum _{i=1}^{M}m{}'_{ij}=\sum _{i=1}^{M}w_{ij}\cdot m_{ij},\qquad s.t. \sum _{i=1}^{M}w_{ij}=1,
\end{aligned}
\label{weight_operation}
\end{equation}
where $R=\left [ r_{1}, ..., r_{j}, ...,r_{C}  \right ], r_{j}\in \mathbb{R}^{W\times H}$ and $m{}'_{ij}\in \mathbb{R}^{W\times H}$ denotes the $j$-th channel of the $i$-th rescaled feature maps $M{}'_{i}$. 
SFFM aims to incorporate hierarchical features with as few parameters as possible and each weight vector $W_{i}, W_{i}\in \mathbb{R}^{C}$ in SFFM depends on global features of all intermediate feature maps, which is different from channel attention in SENets \cite{hu2017squeeze}.

\section{Experiment}
\renewcommand\arraystretch{1}
\begin{table}[bp]
\centering
\caption{ Ablation study of spindle block and SFFM for scale factor $\times 4$ on dataset Urban100}
\begin{tabular}{|c|p{1.5cm}<{\centering}|p{1.5cm}<{\centering}|p{1.5cm}<{\centering}|}
\hline
  &LFFN-NF&LFFN-NS&LFFN\\
\hline
\hline
Spindle Block& \checkmark &\texttimes  & \checkmark \\
SFFM &\texttimes  & \checkmark & \checkmark \\
\hline
Params.(k) & 1,497 &4,770  & 1,531 \\
PSNR/SSIM &25.97/0.7821  &26.20/0.7893 & \textcolor[rgb]{1,0,0}{26.24/0.7902} \\
\hline
\end{tabular}
\label{ablation_study}
\end{table}
\begin{figure}[hb]
\centering
\includegraphics[scale=0.49]{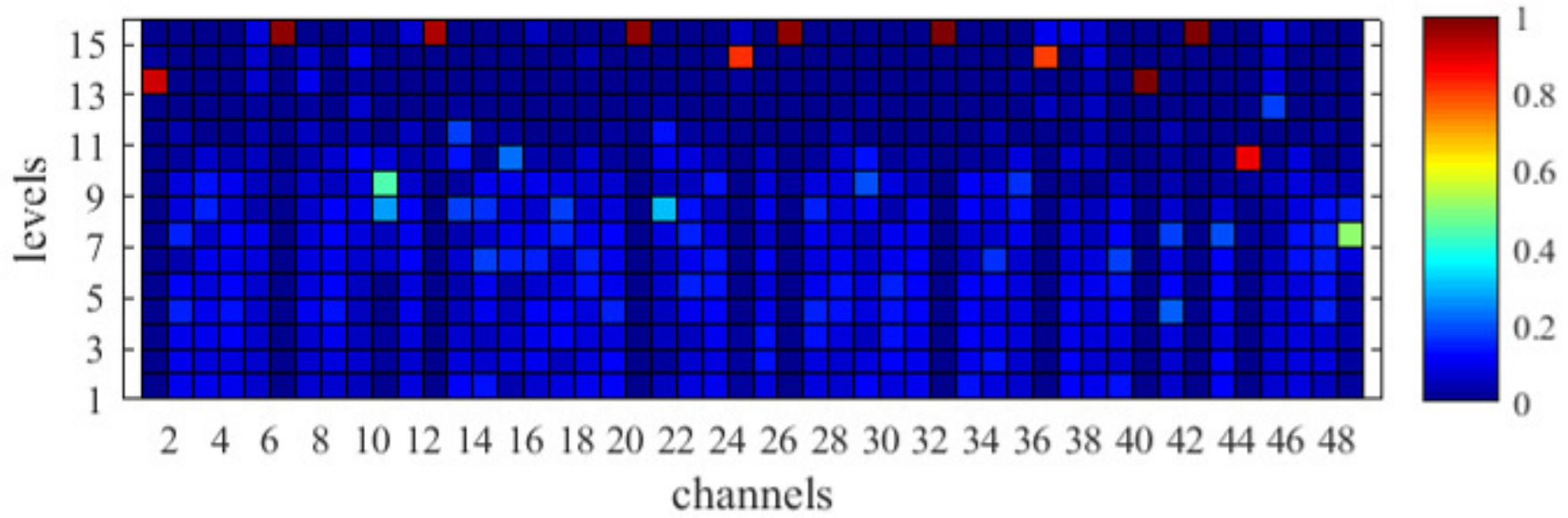}
\caption{Information source distribution for different channels of the feature maps used for reconstruction. The figure is drawn by visualizing the weight vector $W$ of intermediate feature maps for the ``butterfly'' image from Set5 dataset on $\times 4$ enlargement.}
\label{softmax_fuse_dis}
\end{figure}

\subsection{Implementation details}
At first, we pre-train our model on 91 images from Yang \emph{et~al.} \cite{yang2010image} and 200 images from the Berkeley Segmentation Dataset \cite{arbelaez2011contour}. To further improve the performance, we use a newly-proposed high-quality image dataset DIV2K \cite{agustsson2017ntire} which consists of 800 images to fine-tune our pre-trained model. Data augmentation (rotation and flip) is also performed on the 291-image dataset and DIV2K dataset. To produce LR images, we downscale the HR images on particular scaling factors with bicubic interpolation. The proposed method is compared on four widely used benchmark datasets: Set5 \cite{bevilacqua2012low}, Manga109 \cite{matsui2017sketch}, BSD100 \cite{martin2001database}, Urban100 \cite{huang2015single}. For fair comparison, we evaluate the model with PSNR and SSIM on Y channel (i.e., luminance) of transformed YCbCr space.
\par In our final architecture LFFN, 15 spindle modules, each contains 4 spindle blocks, are constructed (i.e., B4M15). We initialize all convolutional filters using the method of He \emph{et~al.} \cite{he2015delving}. We use the L1 loss as our loss function instead of the L2.  For optimization, we use the ADAM optimizer \cite{kingma2014adam} by setting $\beta_{1}=0.9$, $\beta_{2}=0.999$, and $\epsilon=10^{-8}$. We use 16 RGB input patches of size $32\times32$ from the LR images for training, and the initial learning rate is set to $8\times10^{-4}$ and then decreased to half every 20 epochs. In order to accelerate the convergence, we adopt the adjustable gradient clipping \cite{Kim2016Accurate} which has been well implemented in tensorflow. Both training stages are configured the same as demonstrated above except that the initial learning rate is set to $4\times10^{-4}$ during the fine-tuning. Training a LFFN roughly takes four days with a GTX 1080Ti GPU on the $\times$2 model.
\subsection{Model Analysis}
\renewcommand\arraystretch{1}
\begin{table*}[!ht]
\centering
\caption{\small Benchmark SISR results. Average PSNR/SSIM for scale factor $\times$2, $\times$3 and $\times$4 on datasets  Set5, Manga109, BSD100 and Urban100. Red color indicates the best performance.}
\begin{tabular}{|c|c|c|c|cc|cc|cc|cc|}
\hline
    \multirow{2}[2]{*}{Algorithm} & \multirow{2}[2]{*}{scale} & \multirow{2}[2]{*}{ {\ \ Params(K)\ }} & \multirow{2}[2]{*}{Mult-Adds(G)} & \multicolumn{2}{c|}{Set5} & \multicolumn{2}{c|}{Manga109} & \multicolumn{2}{c|}{BSD100} & \multicolumn{2}{c|}{Urban100} \\
          &       &       &       & PSNR  & SSIM  & PSNR  & SSIM  & PSNR  & SSIM  & PSNR  & SSIM \\
\hline
\hline
VDSR \cite{Kim2016Accurate}& 2 & 665 & 612.6 & 37.53& 0.9587& 37.22& 0.9750& 31.90& 0.8960& 30.76& 0.9140\\
DWSR \cite{guo2017deep}& 2 & 373 & 344.0 & 37.42& 0.9568& 37.27 & 0.9719 & 31.85& 0.8944& 30.46& 0.9162\\
DRRN \cite{tai2017image}& 2 & 297 & 6796.9 & 37.74& 0.9591& 37.60& 0.9736& 32.05& 0.8973& 31.23& 0.9188\\
MemNet \cite{tai2017memnet}& 2 & 677 & 2665.0 & 37.78& 0.9597& 37.72& 0.9740& 32.08& 0.8978& 31.31& 0.9195\\
CARN \cite{ahn2018fast}& 2 & 1592 & 222.8 & 37.76& 0.9590& 38.28& 0.9754& 32.09&0.8978& 31.92&  0.9256\\
LFFN & 2 & 1522 & 342.8 & \color{red}37.95&\color{red} 0.9597&\color{red} 38.73& \color{red}0.9765&\color{red} 32.20& \color{red}0.8994& \color{red}32.39&\color{red}0.9299\\
LFFN-S & 2 & 173 & 37.9 & 37.66& 0.9585&37.93& 0.9746& 31.96& 0.8963& 31.28& 0.9192\\
\hline
\hline
VDSR \cite{Kim2016Accurate}& 3 & 665 & 612.6 & 33.66& 0.9213& 32.01& 0.9340& 28.82& 0.7976& 27.14& 0.8279\\
DWSR \cite{guo2017deep}& 3 & 373 & 344.0 & 33.75& 0.9209& 32.14 & 0.9323 & 28.80& 0.7972& 27.22& 0.8293\\
DRRN \cite{tai2017image}&3 & 297 & 6796.9 & 34.03& 0.9244&32.42& 0.9359& 28.95& 0.8004& 27.53& 0.8378\\
MemNet \cite{tai2017memnet}&3 & 677 & 2665.0 & 34.09& 0.9248& 32.51& 0.9369& 28.96& 0.8001& 27.56& 0.8376\\
CARN \cite{ahn2018fast}&3 & 1592 & 118.8 & 34.29& 0.9255&33.47& 0.9429& 29.06&0.8034& 28.06& 0.8493\\
LFFN &3 & 1534 & 153.6 &\color{red} 34.43&\color{red} 0.9266&\color{red} 33.65&\color{red} 0.9445&\color{red} 29.13& \color{red}0.8059&\color{red} 28.34&\color{red} 0.8558\\
LFFN-S & 3 & 185 & 18.1 & 34.04& 0.9233& 32.80&0.9381& 28.91& 0.8005& 27.51& 0.8372\\
\hline
\hline
VDSR \cite{Kim2016Accurate} & 4 & 665 & 612.6 & 31.35& 0.8838& 28.83& 0.8870&27.29& 0.7251& 25.18& 0.7524\\
DWSR \cite{guo2017deep}& 4 & 373 & 344.0 & 31.39& 0.8829& 29.01 & 0.8855 & 27.27& 0.7246& 25.27& 0.7552\\
DRRN \cite{tai2017image}& 4 & 297 & 6796.9 & 31.68& 0.8888&29.18& 0.8914& 27.38& 0.7284& 25.44& 0.7638\\
MemNet \cite{tai2017memnet} & 4 & 677 & 2665.0 & 31.74& 0.8893& 29.42& 0.8942& 27.40& 0.7281& 25.50& 0.7630\\
CARN \cite{ahn2018fast}&4 & 1592 & 90.9 & 32.13& 0.8937&30.45& 0.9073& 27.58&0.7349& 26.07& 0.7837\\
LFFN & 4 & 1531 & 87.9 &\color{red} 32.15&\color{red}0.8945& \color{red}30.66& \color{red} 0.9099&\color{red} 27.52&\color{red}0.7377&\color{red} 26.24& \color{red}0.7902\\
LFFN-S & 4 & 183 & 11.7 & 31.79& 0.8886& 29.76& 0.8979& 27.42& 0.7308& 25.52& 0.7673\\
\hline
\end{tabular}
\label{chart:Benchmark}
\end{table*}
Table \ref{ablation_study} shows the effects of spindle block and softmax feature fusion module (SFFM). LFFN-NF is LFFN without softmax feature fusion module (SFFM) and we replace spindle blocks with residual blocks (Fig.\ref{basic_block}(a)) in LFFN-NS. The three networks have the same number of basic blocks (B4M15). Compared with LFFN, the performance of LFFN-NS degraded and the parameters increased by three times, indicating that the proposed spindle block is more effective than residual block. 
LFFN is obviously superior to LFFN-NF, and the parameters are not increased much, revealing that SFFM is valid for incorporating hierarchical information. Beyond that, as shown in Fig.\ref{softmax_fuse_dis}, the different channel information of the feature maps used for reconstruction come from all levels. And high-level features play a major role in some channels, while low-level features dominate in other channels, which indicates that aggregating hierarchical features is important for SISR and SFFM can implement it well.

\subsection{Comparisons With State-of-the-Art Methods}

We compare LFFN (B4M15) and LFFN-S (B4M4 + depthwise convolution) with state-of-the-art methods. 
We also compare parameters and computation (Mult-Adds) of each method. And Mult-Adds is calculated by assuming that the spatial resolution of HR image is $1280 \times 720$. 
As shown in Table \ref{chart:Benchmark}, our LFFN performs favorably against state-of-the-art methods on all datasets. 
LFFN exceeds Memnet\cite{tai2017memnet} by a margin of 0.41 PSNR while being 30.32 times less compute than Memnet for upscaling $\times 4$ on Set5. Our smallest network LFFN-S has Mult-Adds about $0.44 \%$ of MemNet, $0.17 \%$ of DRRN and $3.40 \%$ of DWSR on $\times 4$ enlargement, respectively, but still achieves comparable performance. 
Fig.\ref{fig:trade_off_time_psnr} shows the execution time of different methods. We use the original codes of state-of-the-art methods to evaluate the runtime on the same machine with 2.1 GHz Intel Xeon CPU and GTX 1080 Ti GPU (12G Memory). LFFN faster, lighter and more accurate than the latest lightweight network CARN \cite{ahn2018fast}. LFFN-S is about 400 times faster and 3.7 times smaller than MemNet

\begin{figure}[t]
\centering
\includegraphics[scale=0.52]{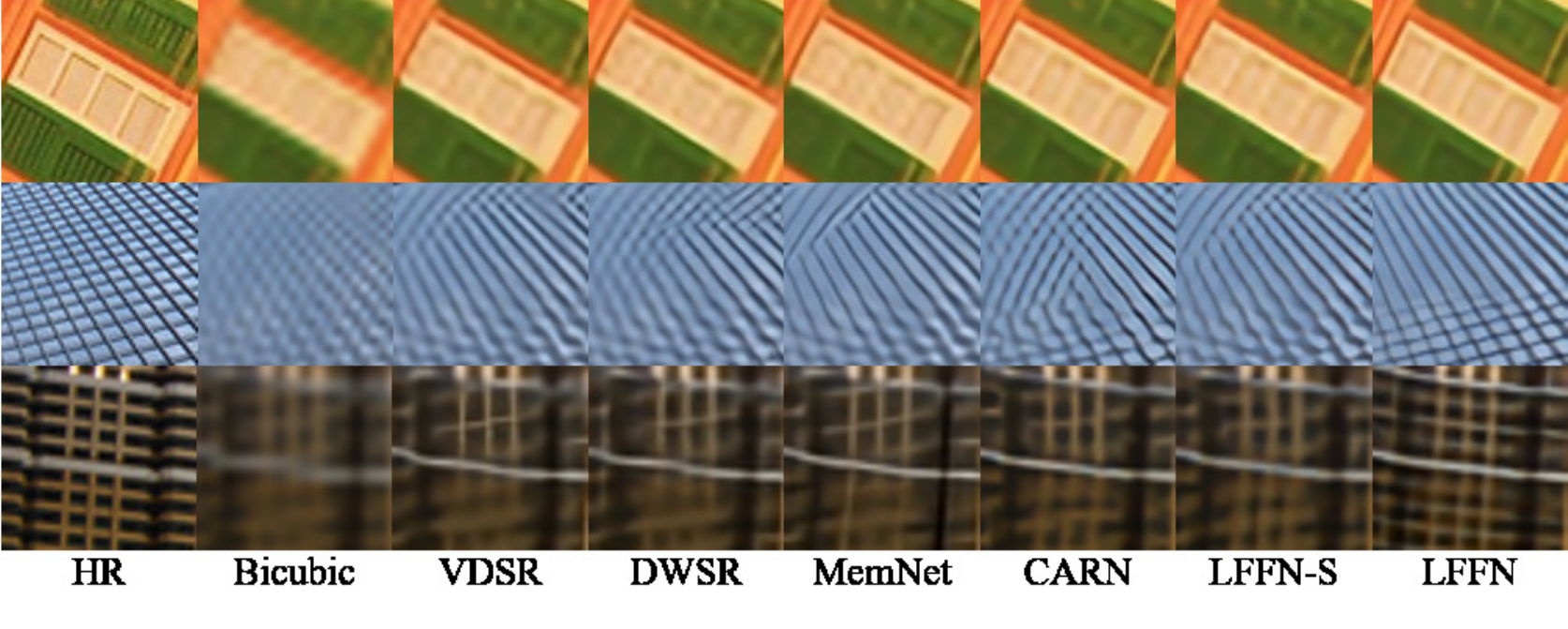}
\caption{Visual comparison for $\times 3$ SR on  ``img013'', ``img062'', ``img085''from the Urban100 dataset.}
\label{visual_compare}
\end{figure}

\par We also provide qualitative comparison in Fig.\ref{visual_compare}. Our smallest network LFFN-S can produce almost the same result as other state-of-the-art methods (e.g., MemNet). Besides, LFFN recovers clearer, more accurate contours and less artifacts than other methods. 

\section{Conclusion}
In this paper,we propose a novel lightweight feature fusion network (LFFN) for single image super-resolution. In order to build a more effective and accurate architecture, we pay more attention to full usage of the feature map information. Whether softmax feature fusion module (SFFM) or the proposed spindle block which serves as the basic building unit can significantly improve the representational capacity of a network with fewer parameters. Experiments well demonstrate the effectiveness of our method.


%



\section*{Acknowledgment}
This work was partly supported by the Natural Science Foundation of China (No.61471216 and No.61771276),  and the Special Foundation for the Development of Strategic Emerging Industries of Shenzhen (No.JCYJ20170817161845824 and No.JCYJ20170307153940960)
 \newpage
\ifCLASSOPTIONcaptionsoff
  \newpage
\fi
\enlargethispage{-10cm}

\ifCLASSOPTIONcaptionsoff
  \newpage
\fi

\bibliographystyle{IEEEtran}
\bibliography{mytex}
\end{document}